\documentclass[lettersize,journal]{IEEEtran}
\usepackage{amsmath,amsfonts}
\usepackage{algorithmic}
\usepackage{algorithm}
\usepackage{array}
\usepackage[caption=false,font=normalsize,labelfont=sf,textfont=sf]{subfig}
\usepackage{textcomp}
\usepackage{stfloats}
\usepackage{url}
\usepackage{verbatim}
\usepackage{graphicx}
\usepackage{cite}
\usepackage{tabularx}  
\usepackage{booktabs}   
\usepackage{multirow}
\usepackage{xcolor}

\begin{document}

\title{From Automated to Autonomous: Hierarchical Agent-native Network Architecture (HANA)}


\author{
    Binghan~Wu,~\IEEEmembership{Member,~IEEE},
    Shoufeng~Wang,~\IEEEmembership{Member,~IEEE},
    Yunxin~Liu,~\IEEEmembership{Fellow,~IEEE},
    Ya\mbox{-}Qin~Zhang,~\IEEEmembership{Fellow,~IEEE},
    Joseph~Sifakis,
    and~Ye~Ouyang,~\IEEEmembership{Fellow,~IEEE}%
\thanks{
    Binghan Wu, Shoufeng Wang, and Ye Ouyang are with 
    AsiaInfo Technologies Limited, Beijing, China 
    (e-mail: \{wubh3, wangsf11, ye.ouyang\}@asiainfo.com).%
}
\thanks{
    Yunxin Liu and Ya-Qin Zhang are with 
    the Institute for AI Industry Research (AIR), Tsinghua University, Beijing, China 
    (e-mail: \{liuyunxin, zhangyaqin\}@air.tsinghua.edu.cn).%
}
\thanks{
    Ye Ouyang is also with the Institute for AI Industry Research (AIR), Tsinghua University, Beijing, China. %
}
\thanks{
    Joseph Sifakis is with Verimag, Université Grenoble Alpes, Grenoble, France 
    (e-mail: joseph.sifakis@univ-grenoble-alpes.fr).%
}
}

\markboth{Journal of \LaTeX\ Class Files,~Vol.~14, No.~8, August~2021}%
{Shell \MakeLowercase{\textit{et al.}}: A Sample Article Using IEEEtran.cls for IEEE Journals}


\maketitle

\begin{abstract}
Realizing Level 4/5 Autonomous Networks (AN) demands a shift from static automation to agent-native intelligence. Current operations, reliant on rigid scripts, lack the cognitive agency to handle off-nominal conditions. To address this, this letter proposes a hierarchical multi-agent reference architecture enabling high-level autonomy. The framework features a Dual-Driven Orchestrator that coordinates specialized Executive Agents, supported by a shared Public Memory for unified domain knowledge. A key innovation is the integration of agent self-awareness, which empowers the system to harmonize deliberative strategic governance with reflexive fault recovery. We instantiate and validate this architecture within a 5G Core environment. Case studies demonstrate that the system sustains critical throughput under congestion and reduces Mean Time to Repair (MTTR) by 86\%, confirming its efficacy in unifying strategic planning with operational resilience.
\end{abstract}


\section{Introduction}

\IEEEPARstart{R}{ealizing} Level 4/5 Autonomous Networks (AN) demands a paradigm shift from static automation to agent-native intelligence. While the industry aims for ``Zero-X" experiences~\cite{chinadaily2025autonomous,TMForum2019AutonomousNetworks}, current human-in-the-loop operations struggle with heterogeneous network complexity. Existing mechanisms (e.g., AIOps~\cite{10247148}, SON~\cite{3GPP_TS_28.313}, SDN orchestration~\cite{ULLAH2025}) act as passive, static controllers. They manage nominal events but lack the cognitive agency to proactively address unforeseen disruptions, leaving a critical gap between automated execution and true autonomous cognition.

\begin{figure*}
    \centering
    \includegraphics[width=\linewidth]{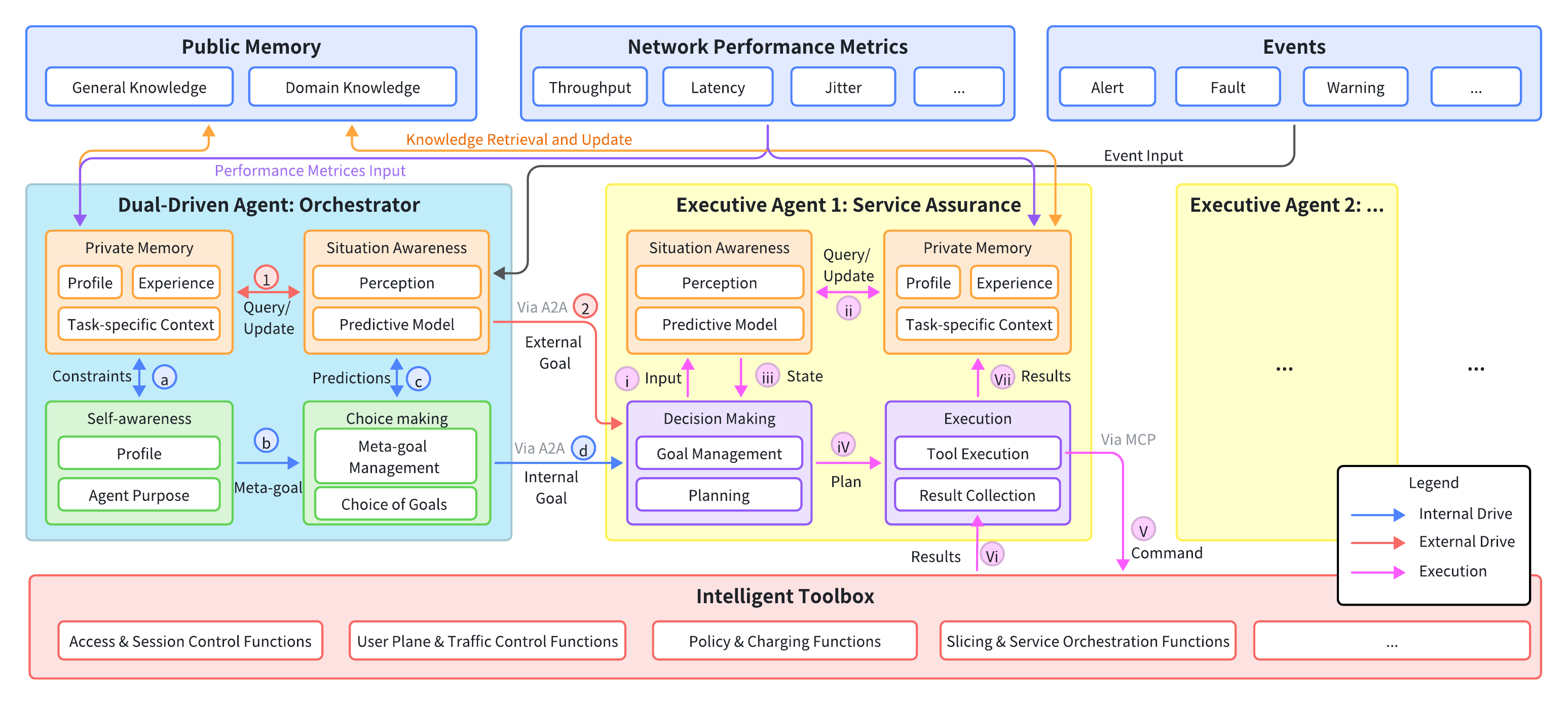}
    \caption{Overview of the proposed Hierarchical Agent-native Network Architecture (HANA). The framework features a Dual-Driven Orchestrator that harmonizes ``slow" Internal Drive (blue flow, a-d) and ``fast" External Drive (red flow, 1-2). It coordinates specialized Executive Agents via the A2A protocol, utilizing a shared Public Memory and an Intelligent Toolbox to achieve a closed-loop perception-cognition-execution cycle for Autonomous Networks.}
    \label{fig:architecture}
\end{figure*}

To bridge this gap, we propose the Hierarchical Agent-native Network Architecture (HANA) (see Figure~\ref{fig:architecture}). Unlike recent intent-driven architectures that merely execute high-level instructions without context awareness~\cite{leivadeas2022survey,li2025agentic}, HANA empowers the network with \textit{intrinsic problem-solving capabilities}. This autonomy is realized by a cognitive architecture grounded in the dual-process theory of ``slow" and ``fast" thinking~\cite{Walter2014Kahneman}:

\begin{itemize}
    \item \textbf{Internal Drive (Slow Thinking):} The system's intrinsic agency is fundamentally derived from an \textit{Internal Drive} akin to ``slow," deliberative cognition, responsible for long-term strategic governance and proactive optimization. In our design, this is operationalized by a \textit{Self-awareness} module that actively maintains internal intent (e.g., service optimality). When it detects a deviation between the current network state and these goals—even in the absence of external faults—it autonomously initiates predictive planning to rectify the trend before performance degrades.
    \item \textbf{External Drive (Fast Thinking):} To complement this deliberation, we superimpose an \textit{External Drive} mimicking ``fast" reactive reflexes for immediate survival. Triggered by critical environmental alerts, this mechanism bypasses the complex reasoning loop, directing Executive Agents to execute pre-validated remedial actions for millisecond-scale fault mitigation.
\end{itemize}

By integrating these two mechanisms, HANA establishes a Dual-Driven Orchestrator Agent that harmonizes long-term strategic governance with immediate operational resilience. This agent operationalizes the dual-process theory through distinct cognitive pathways. For the \textit{Internal Drive}, the Orchestrator interacts with \textit{Long-term Memory} to retrieve system states and constraints, generating an initial meta-goal. This meta-goal is processed by the Decision Making module, which employs predictive cost-benefit analysis to formulate a precise internal goal. Conversely, the \textit{External Drive} responds to external stimuli: the agent synthesizes perception data with the current system context to define an urgent event. Architecturally, HANA achieves a strict decoupling of planning and execution. The Orchestrator functions as the central planner, dispatching the generated goals from the Internal Drive or the External Drive to specialized executive agents. Acting as domain experts, these agents receive the high-level directives and perform localized utility analysis to translate them into concrete, atomic Actions via the Intelligent Toolbox.

The main contributions of this letter are:
(1) We propose a hierarchical, agent-native reference architecture that transitions network management from tool-assisted automation to autonomous problem-solving.
(2) We present a novel dual-driven cognitive framework that decouples strategic cognition from execution, unifying long-term intrinsic intent with short-term surviving need.
(3) We validate HANA in a 5G Core environment. Case studies demonstrate that the system sustains critical throughput under congestion and reduces Mean Time to Repair (MTTR) by 86\% compared to manual O\&M, confirming its efficacy in real-world scenarios.

\section{Hierarchical Agent-native Network Architecture (HANA)}
\label{sec:architecture}

\subsection{Architecture Overview}
As illustrated in Fig. \ref{fig:architecture}, HANA is structured into three logically distinct layers, establishing a closed loop from perception to cognition to execution.

\begin{itemize}
    \item {Public Memory \& knowledge Layer (Top):} This layer acts as the unified knowledge base. It aggregates real-time \textit{Network Performance Metrics} and \textit{Events}. It also maintains a \textit{Public Memory} accessible via the Model Context Protocol (MCP), storing both \textit{General Knowledge} and \textit{Domain Knowledge}. {\color{black} To ensure resilience, the Public Memory adopts a logically centralized but physically distributed design~\cite{3GPPTS23501V1850}. Its lifecycle management—encompassing continuous updating, snapshot-based versioning, and conflict resolution—strictly adheres to industry standards~\cite{ITUTM33512024}.}
    
    \item {Cognitive Core Layer (Middle):} This is the locus of autonomy, hosting the \textit{Dual-Driven Orchestrator Agent} and specialized \textit{Executive Agents} (e.g., Service Assurance Agent). Agents collaborate via the Agent-to-Agent (A2A) protocol to deliberate and decide.
    
    \item {Intelligent Toolbox (Bottom):} This layer bridges cognitive intents to telecom-grade operations. It encapsulates atomic functions (e.g., Access \& Session Control, Policy \& Charging), allowing agents to execute commands safely through standardized interfaces.
\end{itemize}

\subsection{The Dual-Driven Orchestrator: Internal and External Drive}
The core innovation lies in the Orchestrator Agent, which implements a dual-driven cognitive model enabling agents to balance immediate reflexes with long-term strategic planning.

\textbf{The Internal Drive} (blue arrows a--d in Fig.~\ref{fig:architecture}) acts as the strategic planner. Driven by Self-awareness, the agent first retrieves operational constraints and task-specific context from its Private Memory (step~a). It reflects on this internal state to generate a Meta-goal (step~b). {\color{black} The Meta-goal is a persistent, strategic intent stored in Private Memory, and it guides the Choice Making module.} By incorporating forward-looking Predictions derived from situation awareness (step~c), the module formulates high-level strategic objectives. These objectives are finally transmitted as a Internal Goal (step~d) via the A2A protocol. {\color{black} The Internal Goal is a transient, tactical task description.} This extensive reasoning cycle embodies a ``slow thinking'' paradigm that deliberately sacrifices speed to weigh long-term utility and ensure global optimality.

\textbf{The External Drive} (red arrows 1--2 in Fig.~\ref{fig:architecture}) operates in parallel to handle immediate threats. Upon detecting an alert, the Situation Awareness module queries memory layers to retrieve system context and logs the active state (step~1). It then performs a preliminary diagnosis, encapsulating the system state, alert data, and diagnostic results into a Reactive State-Based Event. This event is immediately transmitted via the A2A protocol to a specialized Executive Agent (step~2). This short-circuited pathway enables a ``fast thinking'' reflex that bypasses complex goal reasoning to ensure millisecond-scale mitigation.

\subsection{Executive Agents and Closed-Loop Execution}
Executive Agents translate Orchestrator directives via a rigorous Decision Making process. Upon receiving an input (step i), the agent synthesizes it with historical context (step ii) and real-time state (step iii). The Goal Management module prioritizes tasks. {\color{black} To resolve conflicts between drives, HANA employs a strict priority-based scheduling mechanism. When the External Drive detects a `hard constraint' violation, the resulting event preempts and pauses any ongoing Internal Goals until the fault is mitigated and safe boundaries are restored.} Then, the Planning module generates a concrete Plan (step iv). During Execution, the agent converts this plan into atomic Tool Executions, sending commands to the Intelligent Toolbox via the MCP (step v). Finally, collecting Results (step vi) updates the Private Memory (step vii), ensuring continuous adaptation.

\section{Case Studies}
In this section, we present two distinct case studies to validate the autonomous, closed-loop capabilities of HANA. Case Study A highlights the proactive operation, where the Orchestrator utilizes Self-awareness to anticipate risks and drive strategic optimization. Case Study B demonstrates the reactive mechanism, showcasing how the system executes millisecond-scale self-healing in response to critical faults.

\subsection{Key Terminal Proactive Service Assurance}
This case study validates the architecture's Internal Drive (process a--d and i--Vii in Fig.~\ref{fig:architecture}), focusing on maintaining stringent Service Level Agreements (SLAs) for critical terminals in demanding scenarios, such as industrial IoT and autonomous systems. In these environments, dynamic network congestion is a primary threat. Traditional reactive operations---triggered only after a throughput collapse or service interruption---are fundamentally inadequate, as post-failure mitigation is often too late to prevent mission failure. Therefore, proactive, goal-driven actions that anticipate network risks are necessary. HANA addresses this imperative by leveraging its core cognitive capability, enabling agents to use self-awareness to forecast potential SLA violations and preemptively orchestrate service assurance.

The workflow is initiated by the Orchestrator Agent, responsible for continuously steering network state toward long-term strategic goals. The agent's \textit{Situation Awareness} module (via Perception) continuously ingests telemetry data, specifically monitoring cell-level load trends and correlating them with the presence of high-priority VIP user sessions. Crucially, the agent's \textit{Self-awareness} component actively evaluates these observed, worsening state trends against its maintained \textit{Agent Purpose}---{\color{black} specifically, the meta-goal to ensure a critical application's throughput remains above the 2 Mbps lower bound dynamically retrieved from the VIP terminal's SLA requirement in the Private Memory.} When the \textit{Predictive Model} forecasts a high probability of an imminent SLA violation if no action is taken, the agent's \textit{Choice Making} logic is triggered. Instead of waiting for a failure, it synthesizes the risk prediction with its internal profile to generate a new Meta-goal: ``Execute Preemptive Service Assurance for the VIP Terminal.''

The Orchestrator Agent transmits this objective as a Proactive Goal via the A2A Protocol to the specialized Service Assurance Agent, establishing the initial task context and defining resource guardrails. Upon receiving this goal, the Assurance Agent operates within its proactive behavior loop. Its \textit{Goal Management} and \textit{Planning} modules formulate a comprehensive optimization strategy. This strategy involves preemptively selecting the optimal Next-Generation QoS Identifier (NG-QI), dynamically elevating the critical flow's priority, and reserving the required guaranteed-bitrate allocation to mitigate the impending congestion peak. The agent translates this high-level policy into concrete network configuration changes via the Model Context Protocol (MCP) and the Intelligent Toolbox, ensuring the measure is applied well in advance of the most critical load surge.

\begin{figure}[t]
\centering
\includegraphics[width=.9\linewidth]{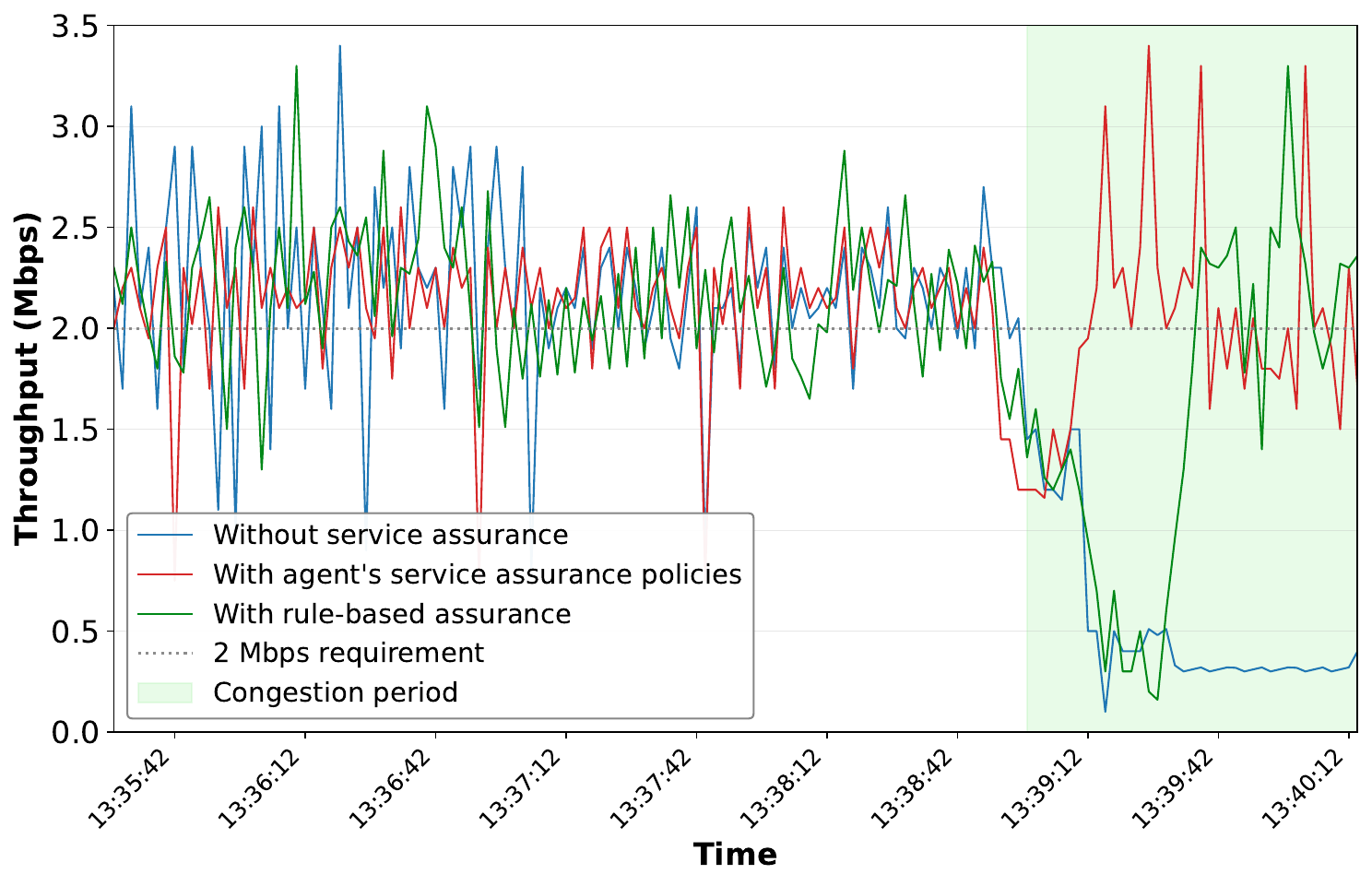}
\caption{\color{black} Video upload rate during congestion for terminals with /without agent-based assurance and traditional rule-based script.}
\label{fig:service}
\end{figure}

{\color{black} To validate this, we simulated a congestion scenario for a 2 Mbps VIP surveillance camera, comparing HANA against an unprotected baseline and a traditional rule-based script. As shown in Fig.~\ref{fig:service}, the unprotected terminal (blue line) suffered a throughput collapse to $\sim$0.25 Mbps. The rule-based script (green line) is inherently reactive; its requisite monitoring debouncing to filter transient fluctuations, and sequential execution introduce a $\sim$30-second latency. Consequently, it intervenes only \textit{after} the SLA is breached, leaving a distinct disruption window. In sharp contrast, HANA (red line) leverages its predictive model to anticipate SLA risks, preemptively orchestrating resource reservations \textit{before} congestion impacts the terminal. This zero-degradation proactive intervention confirms HANA's cognitive superiority over the post-fault mitigation of traditional automation.}

\subsection{Core Network Self-Healing}
This case study validates the architecture's External Drive (process 1--2 and i--Vii in Fig.~\ref{fig:architecture}), addressing the critical need for rapid, automated recovery from hidden core network faults. Modern telecom core networks rely on multi-technology deployments, making them vulnerable to complex issues such as configuration errors or resource exhaustion. Traditional O\&M relies heavily on manual troubleshooting, which is slow, skill-dependent, and inefficient as network scale grows. To minimize the impact of outages on user experience, rapid, autonomous fault recovery has become essential.

The self-healing workflow is triggered when the Orchestrator's \textit{Situation Awareness} module detects a critical anomaly. Consider an instance where the monitoring system raises an ``HTTP Connection Resource Exhaustion'' alarm in a core network Session Management Function (SMF) node. The agent immediately captures this alert, along with relevant network element identifiers and service address details (e.g., 11.12.13.114). In this ``fast thinking'' reflex loop, the agent bypasses complex strategic planning. It immediately queries \textit{Public Memory} (Domain Knowledge) to retrieve historical fault cases and configuration templates. Through pattern matching and contextual analysis, the agent performs a preliminary diagnosis and identifies the root cause: the current maximum HTTP connection setting is 100, significantly lower than the recommended value of 1000 {\color{black} derived by matching historical experience in the Private Memory}.

Instead of triggering a long planning cycle, the Orchestrator encapsulates the system state and this diagnostic result into a Reactive State-Based Event. This event is immediately forwarded via the A2A Protocol to the specialized Self-Healing Executive Agent. Upon receipt, this agent translates the event into a remedial plan: restore normal connection availability without disrupting services. It triggers atomic actions via the Model Context Protocol (MCP) and the \textit{Intelligent Toolbox} to automatically update the maximum HTTP connection parameter to 1000 and perform a graceful configuration reload. The system then continues to monitor connection usage to confirm resolution.


\begin{table}[t]
    \centering
    \footnotesize
    \setlength{\tabcolsep}{3.5pt} 
    \begin{tabularx}{\columnwidth}{|X|c|c|c|c|c|}
        \hline
        \multirow{2}{*}{\textbf{Agent Status}} & \multicolumn{4}{c|}{\textbf{MTTR (min)}} & \multirow{2}{*}{\textbf{Imp(\%)}} \\
        \cline{2-5}
        & \textbf{Dispatch} & \textbf{Analysis} & \textbf{Resolution} & \textbf{Total} & \\
        \hline
        \multicolumn{6}{|c|}{\textbf{Failure: AMF Node Unreachable}} \\
        \hline
        No Agent & 1 & 30 & 5 & 36 & -- \\
        Rule-Based & 1 & 15 & 1 & 17 & 52.78 \\
        With Agent & 1 & 3 & 1 & 5 & 86.11 \\
        \hline
        \multicolumn{6}{|c|}{\textbf{Failure: HTTP Connection Resources Insufficient}} \\
        \hline
        No Agent & 1 & 10 & 3 & 14 & -- \\
        Rule-Based & 1 & 5 & 1 & 7 & 50.00 \\
        With Agent & 1 & 1 & 1 & 3 & 78.57 \\
        \hline
        \multicolumn{6}{|c|}{\textbf{Failure: Total Session Capacity Level-1 Alarm}} \\
        \hline
        No Agent & 1 & 10 & 10 & 21 & -- \\
        Rule-Based & 1 & 5 & 10 & 16 & 23.81 \\
        With Agent & 1 & 1 & 10 & 12 & 42.86 \\
        \hline
    \end{tabularx}
    \vspace{0.1cm}
    \caption{\color{black} MTTR comparison with/without agent and traditional rule-based script.}
    \label{tab:mttr}
\end{table}

To evaluate the agent's self-healing capability, we tested three typical core network failure scenarios:
(1) \textit{Access and Mobility Management Function (AMF) Node Unreachable}, indicating a critical access management function failure affecting user connectivity;
(2) \textit{HTTP Connection Resources Insufficient}, reflecting resource exhaustion in control plane communication;
(3) \textit{Total Session Capacity Level-1 Alarm}, representing service capacity saturation requiring immediate expansion.

{\color{black} Table I evaluates HANA against manual operations (``No Agent") and a rule-based sequential decision tree (e.g., Ping checks $\rightarrow$ Pod status checks $\rightarrow$ Link checks). While accelerating execution, the rule-based script suffers from deterministic traps during analysis: it wastes time on rigid, irrelevant checks before ultimately suspending for human intervention. Conversely, HANA's ``fast thinking" reflex leverages Public Memory for global feature matching, bypassing sequential steps to directly pinpoint root causes. Consequently, HANA slashed AMF fault analysis time to just 3 minutes, compared to 15 minutes for the rule-based script and 30 minutes manually. This confirms HANA elevates ``tool-based automation" to ``cognitive autonomy," effectively eliminating the root cause analysis bottleneck.}

\section{Conclusion and Discussion}
HANA establishes a hierarchical, agent-native architecture for autonomous networks, successfully decoupling strategic cognition from execution. By instantiating a dual-driven model within the Orchestrator, the system harmonizes the ``slow thinking'' required for long-term goal management with the ``fast thinking'' needed for immediate fault resilience. Case studies confirm the architecture's efficacy: sustaining VIP SLAs under congestion through proactive governance and reducing fault repair times by up to 86\% through reactive reflexes. These results validate that embedding self-awareness and distinct cognitive flows is essential for transitioning from automated to truly autonomous networks.

{\color{black} HANA’s hierarchical design is inherently tailored for scalability. Executive Agents can scale horizontally, and each Orchestrator is scoped to its domain to prevent centralized bottlenecks. However, engineering boundaries remain: utilizing Large Language Models (LLMs) introduces inherent inference latency for highly concurrent strategic goals, and achieving seamless cross-domain coordination requires further standardized intent translation mechanisms.} Future work will focus on extending this hierarchical framework to cross-domain scenarios, coordinating autonomous behaviors across Core, RAN, and Transport domains to achieve end-to-end network autonomy.

\bibliographystyle{ieeetr}
\bibliography{reference}

@article{chinadaily2025autonomous,
  author   = {Li Huidi and Ouyang Ye and Joseph Sifakis},
  title    = {Autonomous networks driving the progress of telecom sector},
  journal  = {China Daily},
  date     = {2025-05-24},
  year     = {2025},
  url      = {https://www.chinadaily.com.cn/a/202505/24/WS683115ada310a04af22c1489.html},
  urldate  = {2025-07-09},
  note     = {Updated: 2025-05-24 08:41}
}

@ARTICLE{10247148,
  author={Yang, Yuqian and Yang, Shusen and Zhao, Cong and Xu, Zongben},
  journal={IEEE Communications Magazine}, 
  title={TelOps: AI-Driven Operations and Maintenance for Telecommunication Networks}, 
  year={2024},
  volume={62},
  number={4},
  pages={104-110},
  doi={10.1109/MCOM.003.2300055}}

@techreport{3GPP_TS_28.313,
  author       = {3GPP},
  title        = {Self-Organizing Networks (SON) for 5G networks},
  institution  = {3GPP},
  type         = {Technical Specification},
  number       = {TS 28.313},
  version      = {16.0.0},
  year         = {2020},
  month        = {October},
  url          = {https://www.etsi.org/deliver/etsi_ts/128300_128399/128313/16.00.00_60/ts_128313v160000p.pdf},
  note         = {Release 16},
  language     = {English}
}

@article{ULLAH2025,
title = {Autonomous network management for 6G communication: a comprehensive survey},
journal = {Digital Communications and Networks},
year = {2025},
issn = {2352-8648},
doi = {https://doi.org/10.1016/j.dcan.2025.07.001},
url = {https://www.sciencedirect.com/science/article/pii/S2352864825001129},
author = {Inam Ullah and Ali Arishi and Sushil Kumar Singh and Faisal Alharbi and Anwar Hassan Ibrahim and Muhammad Islam and Yousef Ibrahim Daradkeh and Chang Choi},
}

@article{leivadeas2022survey,
  title={A survey on intent-based networking},
  author={Leivadeas, Aris and Falkner, Matthias},
  journal={IEEE Communications Surveys \& Tutorials},
  volume={25},
  number={1},
  pages={625--655},
  year={2022},
  publisher={IEEE}
}

@article{li2025agentic,
  title={The Agentic-AI Core: An AI-Empowered, Mission-Oriented Core Network for Next-Generation Mobile Telecommunications},
  author={Li, Xu and Shi, Weisen and Zhang, Hang and Peng, Chenghui and Wu, Shaoyun and Tong, Wen},
  journal={Engineering},
  year={2025},
  publisher={Elsevier}
}

@techreport{TMForum2019AutonomousNetworks,
  title        = {Autonomous Networks: Empowering Digital Transformation for the Telecoms Industry},
  author       = {Boasman-Patel, Aaron Richard Earl and Dong, Sun and Wang, Ye and Maitre, ChrisWan and Domingos, José and Troullides, Yiannis and Mas, Ignacio and Traver, Gary and Lupo, Guy},
  institution  = {TM Forum},
  type         = {Whitepaper},
  number       = {Release 1.0},
  address      = {inform.tmforum.org},
  year         = {2019},
  month        = {May},
  day          = {15},
  url          = {https://www.tmforum.org/wp-content/uploads/2019/05/22553-Autonomous-Networks-whitepaper.pdf},
}

@article{Walter2014Kahneman,
  title={Kahneman, D. (2011): Thinking, Fast and Slow},
  author={Walter and Krämer},
  journal={Statistical Papers},
  year={2014},
}

@techreport{3GPPTS23501V1850,
  author = {{3GPP}},
  title = {System architecture for the 5G System (5GS)},
  institution = {3rd Generation Partnership Project (3GPP)},
  type = {Technical Specification (TS)},
  number = {23.501},
  year = {2024},
  month = jun,
  note = {Version 18.5.0},
  url = {https://www.3gpp.org/ftp/Specs/archive/23_series/23.501/}
}

@standard{ITUTM33512024,
  author = {{ITU-T}},
  title = {Framework of knowledge management for telecom operation and management},
  institution = {International Telecommunication Union},
  type = {Technical Specification (TS)},
  number = {M.3351},
  year = {2024},
  month = aug
}
\vfill

\end{document}